# Generalized Belief Propagation on Tree Robust Structured Region Graphs

**Andrew E. Gelfand**
Dept. of Computer Science
University of California, Irvine
Irvine, CA 92697-3425, USA

**Max Welling**
Dept. of Computer Science
University of California, Irvine
Irvine, CA 92697-3425, USA

## Abstract

This paper provides some new guidance in the construction of region graphs for Generalized Belief Propagation (GBP). We connect the problem of choosing the outer regions of a Loop-Structured Region Graph (SRG) to that of finding a fundamental cycle basis of the corresponding Markov network. We also define a new class of *tree-robust* Loop-SRG for which GBP on any induced (spanning) tree of the Markov network, obtained by setting to zero the off-tree interactions, is exact. This class of SRG is then mapped to an equivalent class of *tree-robust* cycle bases on the Markov network. We show that a tree-robust cycle basis can be identified by proving that for every subset of cycles, the graph obtained from the edges that participate in a single cycle only, is multiply connected. Using this we identify two classes of tree-robust cycle bases: planar cycle bases and "star" cycle bases. In experiments we show that tree-robustness can be successfully exploited as a design principle to improve the accuracy and convergence of GBP.

## 1 Introduction

Loopy belief propagation (BP) on Markov networks (MNs) [16, 29, 5, 22, 1, 17] has remained an active area of research for more than a decade. Much progress has been made in characterizing its fixed points [26, 33, 9], improving and understanding its convergence properties [26, 4, 11, 12], designing convergent alternatives [32, 35, 9, 7, 8, 19], developing new message passing algorithms based on alternative convex objectives [27, 28, 10, 25], and extending BP to GBP based on larger clusters (also known as the Kikuchi approximation or Cluster Variation Method) [14, 20, 34].

Despite all this progress, very little attention has been paid to the question of how to actually choose the outer regions (clusters, cliques) for any of the GBP algorithms. This is surprising given that GBP has the potential to improve the accuracy of BP by orders of magnitude without necessarily adding all that much computational burden. Moreover, GBP is very sensitive to the choice of outer regions; a bad choice can lead to poor convergence behavior and little improvement in accuracy, while a good choice can result in fast convergence and an accurate approximation.

Some guidance on choosing regions has appeared in the literature. In [30] a (greedy) region pursuit algorithm was proposed that ranks candidate clusters by sending a limited set of messages through the network. This approach is computationally expensive because all candidate clusters need to be evaluated. Moreover, the algorithm has no way of deciding when to stop adding regions. A similarly expensive "wrapper" approach based on the edge deletion method was published in [2]. A scheme that utilizes mini-bucket elimination to choose regions in a two layer region graph (called a join graph) is presented in [3]. In this region graph, inner regions represent the separators and connections between outer and inner regions are chosen so that the running intersection property holds.

Two desirable properties of region graphs on partially ordered sets (posets) - $k$-connectedness and $k$-balancedness - were proposed in [18, 23]. Those authors also show that the Cluster Variation Method (CVM) [14, 20, 34] yields a region graph that is totally connected and balanced and that the join-graph construction is only 1-connected and 1-balanced. Structured Region Graphs (SRGs) introduced in [31] also inherit the total connectedness and balancedness properties. Another desirable property of region graph constructions discussed in [34] is that the sum of all counting numbers should equal 1. This property ensures exact results for infinitely strong interactions. A condition called maxent-normality, which ensures that the free energy is maximal in the limit of zero interactions, was also introduced in [34]. Maxent-uniqueness was strengthened in [31] to guarantee that uniform beliefs are the unique fixed points of GBP.

In [31] it was shown that Loop-SRGs - a specific class of

SRG with loop outer regions - satisfy all the desirable conditions when the loop regions form a linearly independent set of size #edges - #nodes + 1 - i.e. a cycle basis. Our contribution can be seen as a refinement of the criteria proposed in [31]. In particular, we show that to satisfy the desirable conditions (e.g. non-singularity and sum of counting numbers equal 1), the set of loop regions must form a fundamental cycle basis. We also propose a condition called "tree-robustness" that further restricts the choice of loop regions. The idea of tree-robustness is to require GBP on a SRG to be *exact on every possible tree embedded in the Markov Network*, where an embedded tree is obtained by zeroing out the interactions on the off-tree edges. For loop SRGs, we show that this idea can be translated to an equivalent problem in graph theory, namely that of finding tree-robust cycle bases. We then characterize the class of tree robust cycle bases and identify two families of tree robust cycle bases. Finally, we demonstrate the performance of GBP on Loop-SRGs constructed to satisfy tree-robustness.

## 2 Generalized Belief Propagation and Structured Region Graphs

Belief Propagation (BP) [24] is an algorithm for computing marginal probabilities in distributions taking the following form:

$$p(\mathbf{x}) = \frac{1}{Z} \prod_a f_a(\mathbf{x}_a) \quad (1)$$

where $a$ indexes the factors in the model and $f(\mathbf{x}_a)$ is a function on $\mathbf{x}_a$, which is a subset of $\{x_1, ..., x_n\}$, the $n$ discrete-valued random variables in the distribution. In this paper we assume $p(\mathbf{x})$ to be comprised of only pairwise factors but the extension to factor graphs is straightforward.

BP is a message passing algorithm that computes marginals by iteratively sending messages from one node to its neighboring nodes. It is most easily described in terms of messages passed along the edges of a *factor graph* [15] (factors are pairwise interactions in this paper), which is a bipartite graph comprised of *factor* nodes and *variable* nodes. The edges in a factor graph connect each factor node $a$ to the variable nodes $i$ for which $x_i \in \mathbf{x}_a$. The belief $b_i(x_i)$ at variable node $i$ is an approximation to the exact marginal $p(x_i)$ and is defined as:

$$b_i(x_i) \propto \prod_{a \in N(i)} m_{a \to i}(x_i) \quad (2)$$

where $m_{a \to i}(x_i)$ is a message from factor node $a$ to variable node $i$ and $N(i)$ is the set of factor nodes neighboring variable node $i$. The belief $b_a(\mathbf{x}_a)$ over the variables $\mathbf{x}_a$ is defined in an analogous fashion, as a product of messages from variable nodes to factor node $a$. These beliefs will be exact if the factor graph contains no cycles.

An important limitation of BP is that messages are defined on a single variable. This means that interactions between variables may be lost during message passing. Generalized Belief Propagation (GBP) [34] is a class of algorithms that address this limitation. In GBP, messages over one or more variable are passed among sets of nodes or *regions*. A *region graph* is a structure, analogous to the factor graph, that helps organize the computation of GBP messages.

DEFINITION 1. *A Region $R$ is a set of variable nodes and factor nodes such that if factor node $a$ is in region $R$, all variable nodes $i$ where $x_i \in \mathbf{x}_a$ are also in $R$.*

Let $\mathcal{R}$ denote the set of all regions. Let $\mathbf{x}_R$ denote the set of variables in region $R$ and $f_R(\mathbf{x}_R) = \prod_{a \in R} f_a(\mathbf{x}_a)$ be the factors in region $R$. Every variable and factor must belong to some region.

DEFINITION 2. *A Region Graph (RG) is a directed graph $G(V, E)$, where each vertex $v \in V$ is associated with a region $R \in \mathcal{R}$ and the directed edges $e \in E$ are from some vertex $v_p$ to vertex $v_c$ if $\mathbf{x}_{R_c} \subset \mathbf{x}_{R_p}$, where $\mathbf{x}_{R_i}$ is the set of variables in region $R_i$. In such cases, vertex (region) $p$ is the parent of vertex $c$.*

Let $pa(R)$ denote the parents, $an(R)$ denote the ancestors and $de(R)$ denote the descendants of region $R$. Regions with no parents will be referred to as *outer* regions; all other regions are *inner* regions. The Bethe RG is a RG with outer regions for each factor and inner regions for each variable.

Each region $R$ is associated with a counting number $\kappa_r$ used to define the region-based free energy of a RG. Let $R(i) = \{R \in \mathcal{R} | x_i \in \mathbf{x}_R\}$ denote the set of regions containing variable $i$. A RG is considered *1-balanced* if:

$$\sum_{R \in R(i)} \kappa_R = 1 \; \forall \; i \quad (3)$$

1-balancedness is satisfied if $\kappa_R$ is defined recursively as:

$$\kappa_R = 1 - \sum_{A \in an(R)} \kappa_A \quad (4)$$

A RG is *1-Connected* if the subgraph consisting of the regions $R(i)$ is connected for all $i$. These conditions can be strengthened by considering the set of regions containing a larger set of variables - i.e. $R(s) = \{R \in \mathcal{R} | \mathbf{x}_s \in \mathbf{x}_R\}$ for $\mathbf{x}_s \subseteq \{x_1, ..., x_n\}$. A RG is called *totally* connected and balanced if it is connected and balanced for all sets of variables that are subsets of an outer region - i.e. for any $\mathbf{x}_s \subseteq \mathbf{x}_R$ for outer region $R$ [23]. Junction graphs [18] and Join Graphs [3] are two-layer RG constructions satisfying 1-balancedness and 1-connectednesss, while CVM [14, 21] ensures total connectivity and balancedness.

Many different GBP algorithms can be defined on RGs. The parent-to-child (or canonical) GBP algorithm is one such algorithm. It is defined in terms of messages sent from a parent region to a child region. As in BP, the belief at a particular region $R$ is computed as a product of messages into $R$. However, the messages do not come from the regions neighboring $R$, but rather from regions external to $R$. The belief $b_R(\mathbf{x}_R)$ at region $R$ is given by:

$$b_R(\mathbf{x}_R) \propto f_R(\mathbf{x}_R) \prod_{P \in pa(R)} m_{P \to R}(\mathbf{x}_R) \cdot \quad (5)$$

$$\prod_{D \in de(R)} \prod_{P' \in P'(R)} m_{P' \to D}(\mathbf{x}_D) \quad (6)$$

where $P'(R) = \{pa(D) \setminus \{R \cup de(R)\}\}$. Figure 1 illustrates how beliefs are computed on a 3-by-3 grid.

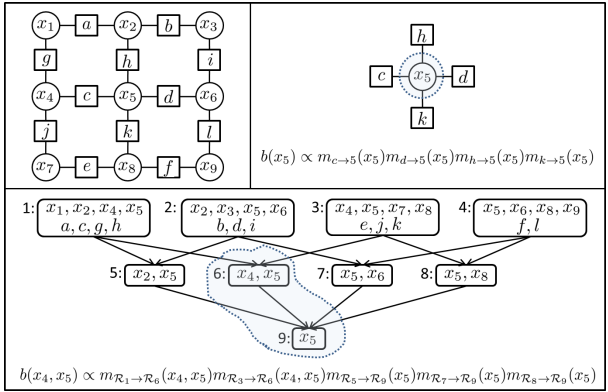

Figure 1: Illustration of message passing on a factor graph (top right) and region graph (bottom) for the 3-by-3 grid in the top-left. In BP, the belief $b_5(x_5)$ is a product of messages into variable node 5. The belief at region 6, $b_{R_6}(x_4, x_5)$, is a product of messages into region 6 and its descendants (i.e. region 9).

The RG framework was extended in [31] to structured message passing algorithms.

DEFINITION 3. *A <u>Structured Region Graph (SRG)</u> is a region graph in which each region $R$ is associated with a set of cliques $\mathcal{C}(R)$. Every variable in $\mathbf{x}_R$ must appear in some clique or factor. The set of factors and cliques associated with a region define a <u>structure</u> $\mathcal{G}(R)$, which is an undirected graph with vertices for each variable in $\mathbf{x}_R$ and edges connecting any pair of variables appearing in the same factor or clique.*

In the rest of the paper, we restrict our attention to Loop-SRGs, which are a particular type of SRG.

DEFINITION 4. *A <u>Loop-SRG</u> is a 3-level SRG consisting of loop outer regions and edge and node inner regions, where a loop outer region has a structure $\mathcal{G}(R)$ that forms an (elementary) cycle. Loop outer regions are connected to the set of edge inner regions comprising the loop and edge inner regions are connected to the two node regions comprising the edge.*

A Loop-SRG on the grid in Figure 1 can be formed by taking the faces of the grid as the loop regions, the edges of the grid as edge regions and all vertices as node regions. Importantly, Loop-SRGs are not limited to planar graphs. In fact, a Loop-SRG can be constructed on any graph structure containing cycles.

## 3 Properties of SRGs

The balancedness and connectedness conditions can be satisfied by many different SRGs for some distribution $p(\mathbf{x})$. However, some SRGs will give better quality approximations than others. Two properties of SRGs that are useful in identifying "good" SRGs were identified in [34]. The first property is that $\sum_R \kappa_R = 1$. This property ensures that the region-based entropy is correct if the variables in $p(\mathbf{x})$ are perfectly correlated. This property will be referred to as *counting number unity*. The second property is *maxent-normality*, which requires the region-based entropy of a connected SRG to achieve its maximum when the region beliefs $b_R(\mathbf{x}_R)$ are uniform. The maxent-normality property was strengthened in [31] to require that message passing with uniform factors have a *unique* fixed point at which $b_R(\mathbf{x}_R)$ are uniform. This stronger property is referred to as *non-singularity*. An SRG that does not satisfy this condition is *singular*.

In [31] it was shown that an acyclic SRG is non-singular. Proving this required a set of reduction operators that modify an SRG's structure while preserving the fixed points of the region-based free energy. The reduction operators will be used in this paper, but are not presented for space reasons (see [31, 30]).

The following qualities of Loop-SRGs come from [31]:

THEOREM 1. *A Loop-SRG has $\sum_R \kappa_R = |L| - |E| + |V|$, where $|L|$ is the number of loop regions, $|E|$ the number of edge regions and $|V|$ the number of node regions.*

THEOREM 2. *A Loop-SRG is singular if $\sum_R \kappa_R > 1$.*

THEOREM 3. *A Loop-SRG is singular iff there is a subset of loop regions and constituent edge regions such that all of the edge regions have 2 or more parents.*

These theorems are illustrated on the grid in Figure 2. In this Loop-SRG, every edge region has exactly 2 loop regions as parents and is thus singular by Theorem 3. There are also $|L| = 3$ loop regions, $|E| = 7$ edge regions and $|V| = 6$ node regions, so that $\sum_R \kappa_R = 2$. Thus, the Loop-SRG is also singular by Theorem 2.

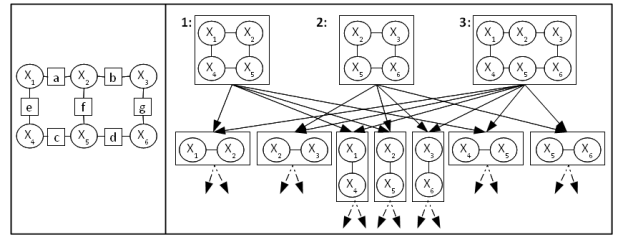

Figure 2: A Loop-SRG (right) for the 2-by-3 grid (left). There are 3 loop outer regions and 7 edge inner regions. The node regions have been dropped for clarity.

Theorems 2 and 3 describe conditions that are undesirable in Loop-SRGs, but offer no guidance on how to identify

"good" Loop-SRGs. The following section establishes a connection between the set of loop regions in a Loop-SRG and cycle bases of a graph that will prove useful in constructing well-behaved Loop-SRGs.

## 4 Loop Regions and Cycle Bases

In [31] it was noted that counting number unity in a Loop-SRG requires $|L| = |E| - |V| + 1$ loops. This number of loops is exactly the dimension of the cycle space of the undirected graph $G$ describing the MN $p(\mathbf{x})$. The connection between loop regions and cycle bases is formalized in this section. We begin with some background on cycle bases taken from [13].

Let $G = (V, E)$ be a 2-connected graph. A simple cycle $C$ in $G$ is a connected Eulerian subgraph in which every vertex has degree 2. The cycle space $\mathfrak{C}(G)$ of a graph $G$ consists of all simple and non-simple cycles of $G$ including the empty cycle $\emptyset$. The dimension of a cycle space for a graph with 1 connected component is $\mu \equiv \mu(G) = |E| - |V| + 1$.

DEFINITION 5. *A Cycle Basis of $\mathfrak{C}(G)$ is a set of simple cycles $\mathcal{B} = \{C_1, ..., C_\mu\}$ such that for every cycle $C$ of $G$, there exists a unique subset $\mathcal{B}_C \subseteq \mathcal{B}$ such that the set of edges appearing an odd number of times in $\mathcal{B}_C$ comprise the cycle $C$.*

DEFINITION 6. *A cycle basis $\mathcal{B}$ is a Fundamental Cycle Basis (FCB) if there exists a permutation $\pi$ of the cycles in $\mathcal{B}$ such that $C_{\pi(i)} \setminus \{C_{\pi(1)} \cup \cdots \cup C_{\pi(i-1)}\} \neq \emptyset$ for $i = 2...\mu$. In other words, the cycles can be ordered so that cycle $C_{\pi(i)}$ has some edge that does not appear in any cycle preceding it in the ordering.*

Consider mapping each loop outer region $R$ to a cycle $C$ in the basis $\mathcal{B}$ of the undirected graph of $p(\mathbf{x})$.[1] Using this mapping, we can claim the following.

THEOREM 4. *A Loop-SRG is Non-Singular and satisfies Counting Number Unity if its loop outer regions are a Fundamental Cycle Basis (FCB) of $G$.*

The proof[2] uses reduction operators to show that a loop outer region with a unique edge (i.e. an edge not shared with any other loop region) can be *reduced* to the set of edges comprising that loop. Since the set of loops form a FCB, we are guaranteed to find a loop region with a unique edge if we reduce loops along $\pi$ - i.e. beginning with the loop corresponding to $C_{\pi(\mu)}$ and ending at loop $C_{\pi(1)}$.

This result implies that the loop regions in an SRG should form a FCB. This greatly reduces the set of loops considered when constructing a Loop-SRG. However, a graph may have many fundamental bases, so one may ask if Loop-SRGs formed from certain FCBs are better than others. The next section defines a class of Loop-SRG with loop regions corresponding to a specific type of FCB.

## 5 Tree Robust Cycle Bases

Non-singularity and counting number unity ensure that GBP behaves sensibly in two opposite and extreme situations. Non-singularity requires GBP to be exact in the presence of uniform factors, while counting number unity ensures that GBP is exact in the presence of perfectly correlated variables.

*Tree-Robustness* is a condition which ensures that GBP behaves sensibly in an orthogonal way. The idea of *Tree-Robustness* is as follows. Imagine removing some set of factors from $p(\mathbf{x})$ to produce a modified MN $p'(\mathbf{x})$ that is acyclic. It would be nice if GBP run on the original Loop-SRG of $p(\mathbf{x})$ but retaining only the factors of $p'(\mathbf{x})$, were equivalent to BP run directly on the Bethe RG for $p'(\mathbf{x})$ (since BP will be exact in that case). A Tree Robust SRG is a Loop-SRG that is exact w.r.t. all acyclic MNs produced by retaining some of the factors in $p(\mathbf{x})$.

More formally, let $T$ be some spanning tree of the undirected graph $G$ describing $p(\mathbf{x})$ and let $\bar{T}$ denote the off-tree edges. A SRG is Tree Exact w.r.t. to $T$ if by removing only factors on edges $\bar{T}$, the SRG can be *reduced* so that inference on $T$ is exact. In other words, a sequence of reduction operations can be applied to the SRG so that it can be reduced to a Bethe RG with edge regions for each edge in $T$ and node regions for each variable. Since $T$ is acyclic, running GBP on this resulting RG will be exact. Finally, we say that a SRG is Tree Robust (TR) if it is Tree Exact w.r.t. all spanning trees of $G$.

The previous section established that a Loop-SRG is non-singular if its loops form a FCB. In this section, we first define a TR cycle basis and then show that a Loop-SRG is TR if its loops form a TR cycle basis.

DEFINITION 7. *Let $T$ be some spanning tree of $G$. A cycle basis $\mathcal{B}$ is Tree Exact w.r.t. $T$ if there exists an ordering $\pi$ of the cycles in $\mathcal{B}$ such that $\{C_{\pi(i)} \setminus \{C_{\pi(1)} \cup \cdots \cup C_{\pi(i-1)}\}\} \setminus T \neq \emptyset$ for $i = 2...\mu$. In other words, the cycles can be ordered so that cycle $C_{\pi(i)}$ has some edge that: 1) does not appear in any cycle preceding it in the ordering; and 2) does not appear in the spanning tree $T$.*

DEFINITION 8. *A cycle basis $\mathcal{B}$ is Tree Robust (TR) if it is Tree Exact w.r.t. all spanning trees of $G$.*

By mapping each loop outer region to a cycle in basis $\mathcal{B}$ it can be shown that[3]:

---

[1] More specifically, the structure $\mathcal{G}(R)$ of each outer region $R$ is chosen to be a unique cycle $C \in \mathcal{B}$

[2] Formal proofs of all theorems in this paper are provided as supplementary material.

[3] See supplementary material for formal proof.

THEOREM 5. *A Loop-SRG is Tree Robust if its loop outer regions are a Tree Robust cycle basis of $G$.*

In the remainder of this section, we introduce two theorems that characterize TR cycle bases. These results prove useful in showing that, for example, the faces of a planar graph constitute a TR cycle basis. The ensuing theorems require the following definition.

DEFINITION 9. *The Unique Edge Graph of a set of cycles $\mathcal{C} = \{C_1, ..., C_k\}$ is a graph comprised of the set of edges that are in exactly one cycle in $\mathcal{C}$. We will use $I(\mathcal{C})$ to denote the unique edge graph for cycle set $\mathcal{C}$. $I(\mathcal{C})$ is cyclic if it contains at least one cycle.*

Using this definition we can now state the following equivalent characterization of TR cycle bases.

THEOREM 6. *Let $\mathfrak{B}^{|k|}$ denote all size $k$ subsets of cycles in $\mathcal{B}$. A FCB $\mathcal{B}$ is Tree Robust iff $I(\mathcal{B}_k)$ is cyclic and not empty for all $\mathcal{B}_k \in \mathfrak{B}^{|k|}$ for $1 \leq k \leq \mu$. In other words, the unique edge graph must be cyclic for all pairs of cycles, and all triples of cycles,..., and all of the $\mu$ cycles.*

COROLLARY 1. *An FCB is TR iff $\mathcal{B}_k$ is TR for all $\mathcal{B}_k \in \mathfrak{B}^{|k|}$ for $1 \leq k \leq \mu$.*

We sketch the main idea of the proof here. Sufficiency follows because whatever ordering $\pi$ we choose to remove the cycles in, there is never a tree $T$ that can block all of the edges of the unique edge graph under consideration (since it is always cyclic). Necessity follows because if there were a subset of cycles for which the exposed edge graph is acyclic then there exists no ordering $\pi$ that can avoid that subset (or a larger subset with an acyclic unique edge graph). Hence, by choosing the tree $T$ to block all edges of the acyclic unique edge graph under consideration we prove that the bases is not tree robust.

## 6 Planar and Complete Graphs

Using the theorems from the previous section, we now identify TR cycle bases of planar and complete graphs. In the case of planar graphs, a TR basis can be constructed from the cycles forming the faces of the planar graph. This supports the observation of previous authors that GBP run on the faces of planar graphs gives accurate results.

THEOREM 7. *Consider a planar graph $G$. The cycle basis $\mathcal{B}$ comprised of the faces of $G$ is TR.*

*Proof.* We use theorem 6. Consider the graph formed by any subset of $k$ faces of the planar graph. Consider any of its connected components. The path that traces the circumference of that component is a loop and also consists of unique edges. □

In the case of complete graphs, a TR basis can be constructed by choosing some vertex as a root, creating a spanning tree with edges emanating from that root and constructing loops of length 3 using each edge not in the spanning tree. This is exactly the 'star' construction of [31], which was shown empirically to be superior to other Loop-SRGs on complete graphs.

THEOREM 8. *Consider a complete graph $G$ on $n$ vertices (i.e. $K_n$). Construct a cycle basis $\mathcal{B}$ as follows. Choose some vertex $v$ as the root. Create a 'star' spanning tree rooted at $v$ (i.e. with all edges $v$-$u$). Now construct cycles of the form $v$-$i$-$j$ from each off-tree edge $i$-$j$. The basis $\mathcal{B}$ constructed in this way is TR.*

*Proof.* We use again theorem 6. Consider the graph constructed from any subset of $k$ triangles. The edges not on the spanning tree (i.e. not connecting to the root) are all unique and will either form a loop (in which case we are done) or a tree. In case of a tree, consider a path connecting two leaf nodes, which are both also connected to the root, thus forming a loop. Because the two edges connecting to the root are also unique (since they correspond to leaf nodes) we have proven the existence of a cycle in the unique edge graph. □

This result can be extended to partially complete graphs where there exists some vertex $v$ that is connected to all the other vertices in $G$.

## 7 TR SRGs in General Graphs

The previous section identified TR cycle bases for two specific classes of graphs. The prescription for how to construct TR SRGs for MNs on these types of graphs is clear: simply find a TR basis $\mathcal{B}$ of the underlying graph and make the cycles in $\mathcal{B}$ the loop regions. This prescription can be generalized in some cases to graphs containing many TR components. More formally,

THEOREM 9. *Consider a graph $G$ comprised of components (subgraphs) $H_1, ..., H_k$. Let the components $H_1, ..., H_k$ be mutually singly connected if for any two components $H_i$ and $H_j$ there exist vertices $v_i \in H_i$ and $v_j \in H_j$ that are singly connected (i.e. connected through a single path). Let $\mathcal{B}_{H_1}, ..., \mathcal{B}_{H_k}$ denote the TR cycle bases for each component. Then a TR cycle basis of $G$ is $\mathcal{B}_G = \{\mathcal{B}_{H_1} \cup \cdots \cup \mathcal{B}_{H_k}\}$.*

*Proof.* First note that by singly connecting the components $H_1, ..., H_k$ we do not create any new cycles. Thus $\mathcal{B}_G$ is a cycle basis of $G$. Every subset of cycles of $\mathcal{B}_G$ is the union of some subset of cycles from the component cycle bases $\{\mathcal{B}_{H_i}\}$. Moreover, for any of these subsets the unique edge graph must be cyclic (by theorem 6). Since the component cycle bases do not overlap it thus follows that the unique edge graph for every subset of cycles of $\mathcal{B}_G$ must also be cyclic. □

For MNs defined over more general graphs, the picture of how to construct a TR SRG is less clear. Since verifying that a basis is TR requires inspecting all subsets of cycles in that basis, searching for a TR basis in a general graph seems difficult. Moreover, not every graph will admit a TR basis. In this section we describe a method for constructing Loop-SRGs that are partially TR. In other words, we sacrifice finding a TR basis of $G$ to find a basis that is Tree Exact for many (just not all) spanning trees of $G$.

The method for finding a partially TR basis works as follows. We first find the largest complete or planar subgraph $H$ of $G$ and construct a TR basis $\mathcal{B}(\mathcal{H})$ for $H$ as described in the previous section. Since the TR core $H$ is a subgraph of $G$, the $\mu(H)$ cycles in $\mathcal{B}(\mathcal{H})$ will not form a complete basis of $G$. We choose the remaining $\mu(G) - \mu(H)$ cycles so that the basis of $G$ is fundamental. We do so by finding a sequence of *ears* (simple paths or cycles) in $G$, such that each new *ear* has some edge not occurring in some previous ear. This process is described in Algorithm *Construct Basis* and is illustrated on a $2 \times 3$ grid in Figure 3.

---
**Algorithm: Construct Basis**

---
**Input**: MN $p(\mathbf{x})$ described by undirected graph $G$
**Output**: Cycle Basis $\mathcal{B}$
    Find TR subgraph $H$ of $G$ with TR basis $\mathcal{B}(\mathcal{H})$
    **if** $G$ contains no TR subgraph (i.e. $H = \emptyset$) **then**
        Let $H$ be some simple cycle in $G$
    **end if**
    Add cycles $\mathcal{B}(\mathcal{H})$ to $\mathcal{B}$
    Mark all edges in $H$ as *used*
    Mark all vertices in $H$ as *visited*
    **while** $\exists$ *unused* edge $e = (s,t)$ from a *visited* vertex $s$ **do**
        If $t$ is *visited*, then set $p_1 = e$
        Else, find an ear $p_1$ from $s$ through edge $e = (s,t)$ to some *visited* vertex $u$.
        Find shortest path $p_2$ from $s$ to $u$ on *used* edges
        Add cycle $C$ consisting of $p_1 \cup p_2$ to the bases $\mathcal{B}$
        Mark all edges (vertices) on $C$ as *used* (*visited*)
    **end while**

---

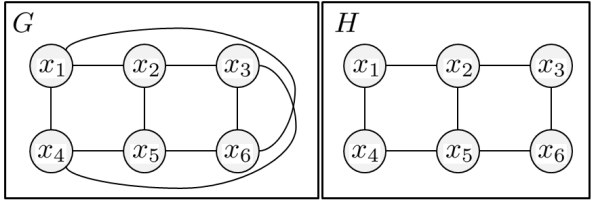

Figure 3: A graph $G$ (left) and its TR core $H$ (right). Add the faces $(1,2,5,4)$ and $(2,3,6,5)$ of $H$ to $\mathcal{B}$. Mark all edges (vertices) in $H$ as used (visited). Edge $(1,6)$ is an ear (unused edge). The path $6, 3, 2, 1$ connects the start and end vertices of this ear along used edges, so cycle $(1,6,3,2)$ is added to $\mathcal{B}$. Similarly, cycle $(3,4,5,6)$ might be added for ear $(3,4)$, giving us the partially TR basis $\mathcal{B} = \{(1,2,5,4), (2,3,6,5), (1,6,3,2), (3,4,5,6)\}$.

## 8 Experiments

We conducted a series of experiments to validate the recommendations for constructing Loop-SRGs made in this paper. In each of the experiments that follow, we generated a set of $M$ random MNs with binary variables. Each MN instance has unary potentials of the form $f_i(x_i) = [\exp(h_i); \exp(-h_i)]$ and pairwise potentials of the form $f_{ij}(x_i, x_j) = [\exp(w_{ij}) \exp(-w_{ij}); \exp(-w_{ij}) \exp(w_{ij})]$. The values of $h_i$ and $w_{ij}$ were drawn from Normal distributions $\mathcal{N}(0, \sigma_{h_i}^2)$ and $\mathcal{N}(0, \sigma_{w_{ij}}^2)$, respectively. Two different error measures are reported. $Error_Z = |\log Z - \log \tilde{Z}|$ is the absolute error in the exact ($Z$) and approximate ($\tilde{Z}$) values of the log partition function. $Error_{L_1}$ measures the error in the marginal probabilities and is computed as $Error_{L_1} = \frac{1}{n} \sum_{i=1}^{n} |b(x_i) - p(x_i)|$ where $n$ is number of variables in a MN instance. Averages of $Error_Z$ and $Error_{L_1}$ were computed across the $M$ MN instances. Error bars indicate standard error. To aid convergence, GBP was run with a damping factor of $0.5$ for at most $1000$ iterations.

### 8.1 TR Bases vs non-TR Bases

We first ran a set of experiments to show that tree robustness is a desirable property. To test this hypothesis, we generated 31 different Loop-SRGs for a MN defined on a complete graph with 20 binary variables (i.e. on $K_{20}$). We first constructed a TR basis $\mathcal{B}$ comprised of all cycles of length 3 passing through vertex 1 (e.g. using the *star* construction from Section 6 with vertex 1 as root). From this TR basis, a sequence of 30 fundamental bases were created as follows: for $i = 1...30$, we choose a cycle $C_i$ of the form $C_i = (1, u, v)$ from $\mathcal{B}$ and modify it by swapping vertex 1 with some vertex $w$ ($w \neq u \neq v \neq 1$) so that $C_i = (w, u, v)$. At every iteration we choose cycles that have not been modified and reject modifications that make the basis non-fundamental. In this way, as $i$ increases the basis is made less TR but always remains fundamental.

Figure 4 shows $Error_{L_1}$ and $Error_Z$ as the Loop-SRG is made less TR. In this figure, we generated $M = 500$ random MN instances with $\sigma_{h_i} = 1$ and $\sigma_{w_{ij}} = 1/\sqrt{20-1}$. Note that by keeping the cycle length at 3 and ensuring that each basis is fundamental, the increase in error can only be explained by the change in the TR core of the basis. Since error increases as the basis is made less TR, these results support the hypothesis that tree robustness is a desirable property. It was also observed that GBP took an increasing number of iterations to converge as the basis was made less TR.

We also ran a set of experiments to characterize the convergence properties of TR SRGs on Ising grid models. Though not reported, these experiments confirm the finding in [34] that running GBP on an RG where the outer regions are

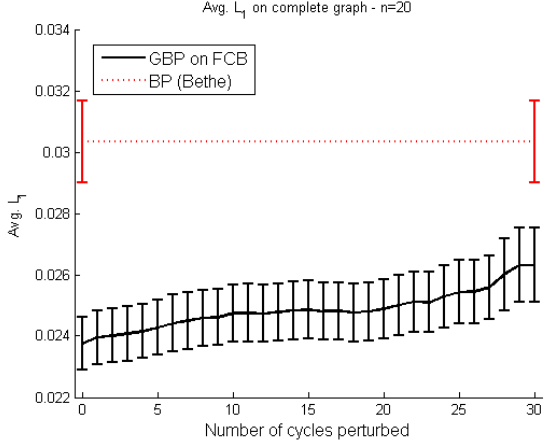

(a) $Error_{L_1}$ on increasingly non-TR Loop SRGs.

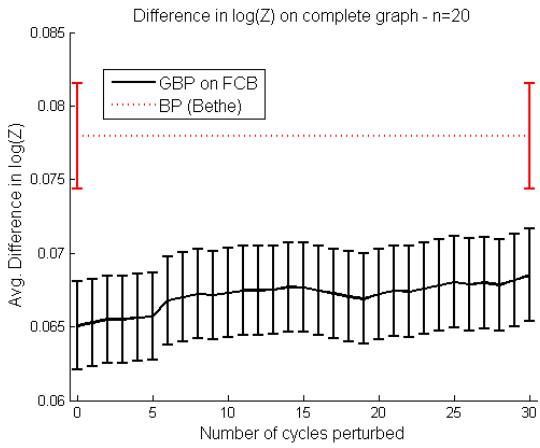

(b) $Error_Z$ on increasingly non-TR Loop SRGs.

Figure 4: Performance of GBP run on a sequence of increasingly non-TR Loop-SRGs.

taken as the faces of the grid model is more accurate than ordinary BP. We also found the TR SRG construction to be more accurate than GBP run on Loop-SRGs constructed from fundamental, but non-TR cycle bases on Ising grids.

## 8.2 Partially TR SRGs

The previous experiment considered a class of MNs for which a TR basis is known. This section considers more general MNs.

The algorithm in Section 7 seeks an initial TR subgraph $H$ of the graph $G$. The previous experiments showed that GBP yields accurate approximations when $H = G$. When a graph contains no TR core (i.e. $H = \emptyset$) the algorithm *Construct Basis* method simply finds a fundamental basis of $G$. We wish to study the performance of GBP between these two extremes - i.e. on partially TR SRGs. We conducted experiments on two types of MNs, where the size of $H$ (relative to $G$) can be controlled.

The following experiments include a comparison to the Iterative Join Graph Propagation (IJGP) method [3]. This method forms a join graph (i.e. a 1-connected and 1-balanced RG) using a heuristic, "mini-bucket" clustering strategy. In IJGP, the number of variables appearing in an outer region is restricted to be less than or equal to an *iBound* parameter. The *iBound* controls the computational complexity of message computations in IJGP because in the join graph construction each outer region forms a clique over all variables in that region. Importantly, since Loop-SRGs assume a loop structure in the outer regions, message computations on Loop-SRGs are equivalent to IJGP with $iBound = 3$.

### 8.2.1 Partial $K$-Trees

In these experiments we construct a set of partial $K$-tree instances via the following procedure[4]. We first build a random $K$-tree on $n$ vertices using the process described in [6]. The number of neighbors (degree) of the vertices in $K$-trees constructed by this procedure follow a power law. This means there will exist a few vertices that are adjacent to most of the vertices in $G$. As a result, the TR core will comprise a large portion of $G$. To reduce the size of the TR core, we iteratively remove edges from the $K$-tree as follows: choose a vertex $v$ with probability proportional to the current degree of that vertex. Modify $G$ by removing an edge from $v$ to one of its neighbors, so long as removing that edge does not disconnect $G$. This process is repeated until the ratio of the maximum degree in $G$ to $n$ falls below some threshold. We refer to this ratio as the connectivity of the graph. A random MN is formed over each partial $K$-tree structure by assigning random unary and pairwise potentials to the vertices and edges.

Figure 5 shows the performance of GBP as a function of connectivity. In these figures, we generated $M = 100$ random MN instances at each level of connectivity (with $K = 10$, $n = 100$, $\sigma_{h_i} = 1$ and $\sigma_{w_{ij}} = 0.3$). The partially TR SRGs are found by choosing the TR core to be the subgraph $H$ found using the max degree vertex as the root of the *star* construction described in Section 6. Cycles are added to this TR core as described in algorithm *Construct Basis*. The partially TR SRGs are compared to Loop-SRGs formed by finding a fundamental cycle basis (FCB) of each partial $K$-tree (constructed using algorithm "*Construct Basis*" with $H = \emptyset$). Importantly, these FCBs do not build upon the TR core. Figure 5 shows that the benefit of the partially TR SRG diminishes as connectivity is decreased. This behavior confirms our belief that the benefit of finding a TR core decreases as the TR core comprises a smaller proportion of cycles in the fundamental basis. Even so, it is important to note that choosing a Loop-SRG with outer regions forming a FCB yields more accurate approximations than both IJGP and BP.

---

[4]$K$-trees are chordal graphs w/ size $K$ maximal cliques. Partial $K$-trees are non-chordal graphs w/ size $K$ maximal cliques

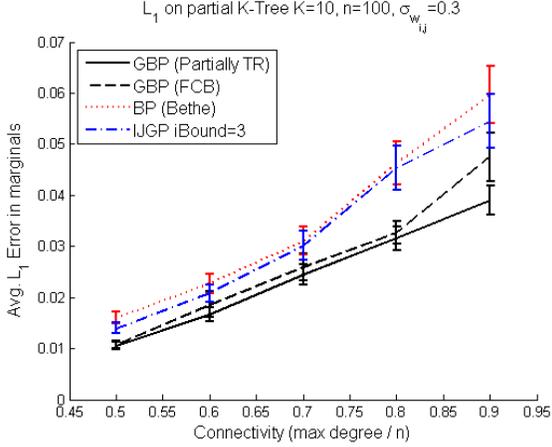

(a) $Error_{L_1}$ as a function of *connectivity*.

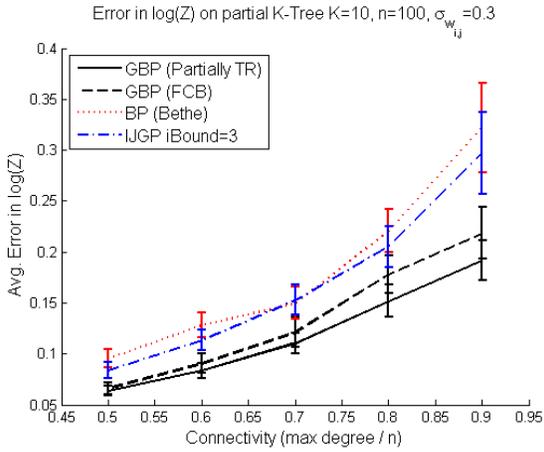

(b) $Error_Z$ as a function of *connectivity*.

Figure 5: Performance of GBP run on the partial TR construction, the FCB construction and IJGP with $iBound = 3$ as a function of *connectivity* (see text for discussion).

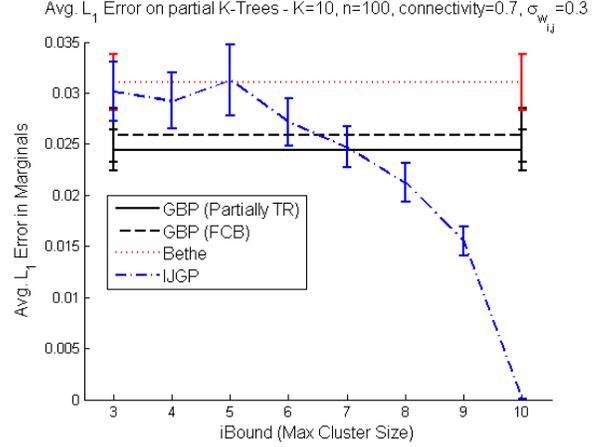

(a) $Error_{L_1}$ as a function of $iBound$.

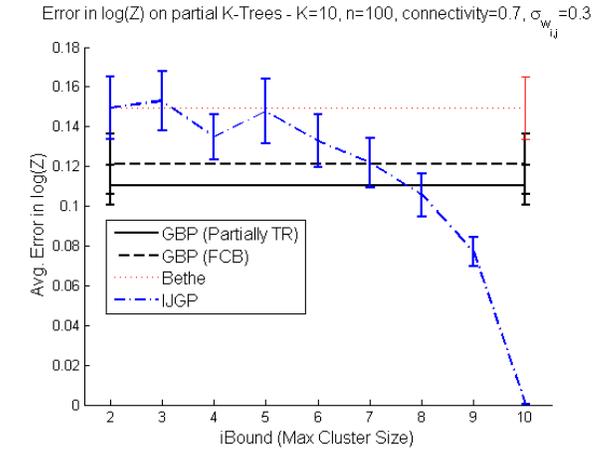

(b) $Error_Z$ as a function of $iBound$.

Figure 6: Performance of the SRG constructions as a function of $iBound$ with fixed *connectivity* of 0.7.

Figure 6 shows the performance of GBP as a function of $iBound$ for a fixed connectivity level. $iBound$ is increased from 2 (which is equivalent to BP) to 10 (which is exact). These plots show that GBP run on the partial TR SRG is roughly equivalent to IJGP with $iBound = 7$, while GBP run on the FCB is equivalent to IJGP with $iBound = 6$. The fact that GBP run on both SRGs performs better than IJGP with $iBound = 6$ is quite remarkable considering that message passing on Loop-SRGs is computationally equivalent to IJGP with $iBound = 3$.

### 8.2.2 Grids with long range interactions

In addition to the partial $K$-tree instances, we also considered grid instances with an increasing number of long range interactions. The additional interactions were added via the following procedure. We begin with a $10 \times 10$ grid. Let $G_0$ denote this initial graph. Two vertices $u$ and $v$ are randomly chosen from the grid. If edge $(u, v)$ exists in graph $G_{i-1}$, new vertices $u$ and $v$ are chosen randomly; if edge $(u, v)$ is not in $G_{i-1}$, then graph $G_i$ is created by adding edge $(u, v)$ to $G_{i-1}$. This process is repeated until a specified number of edges have been added to $G_0$.

Figure 7 compares the $Error_{L_1}$ and $Error_Z$ of GBP run on different loop SRG constructions. $M = 25$ instances were generated with $5, 10, ...,$ and 50 additional edges. In all 250 of these MN instances the unary and pairwise terms were drawn with $\sigma_{h_i} = 1$ and $\sigma_{w_{ij}} = 0.5$. For the partially TR SRG construction, we take the TR core $H$ to be $G_0$ and fill out the cycle basis using algorithm *Construct Basis* (as illustrated in Figure 3). As in the partial $K$-tree experiments, for the FCB construction we choose a fundamental basis that does not build upon the TR core by using algorithm "*Construct Basis*" with $H = \emptyset$. In Figure 7, we see that both the partially TR and FCB construction outperform IJGP with an equivalent $iBound$. Interestingly, when adding 50 additional edges we do not see the $Error_{L_1}$ of the partially TR and the FCB constructions coalesce. This may be explained by the fact that even with 50 additional edges more than 60% of the loops in the SRG are TR (131 cycles in the basis, 81 of which come from the TR core).

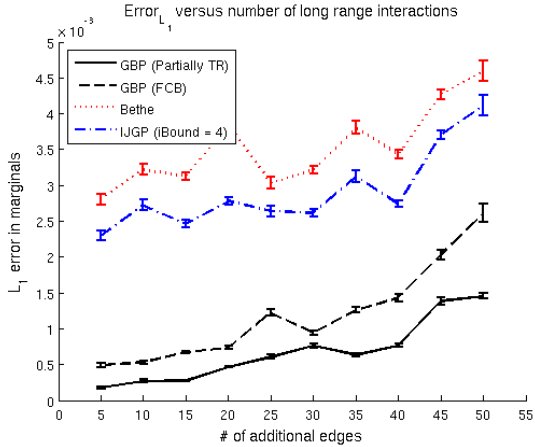

(a) $Error_{L_1}$ on grids with long range interactions

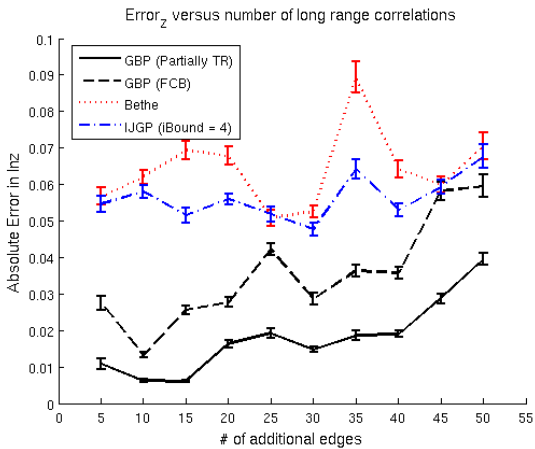

(b) $Error_Z$ on grids with long range interactions

Figure 7: Performance of the different SRG constructions as an increasing number of long range interactions are added to a $10 \times 10$ grid.

## 9 Conclusion

This paper provides some new guidance in the construction of region graphs. In particular, we connected the problem of choosing the loop outer regions of a Loop-SRG to that of finding a fundamental cycle basis of the undirected graph describing a MN. We proposed tree robustness as a refinement to the criterion that the loop regions form a fundamental basis and offered a characterization of the class of TR cycle bases. We identified TR cycle bases for planar and complete graphs. This characterization helps explain the success of GBP on the "star" construction of [31] for complete graphs and the "all faces" construction on planar graphs. We also proposed a practical Loop-SRG construction that first identifies a TR core and then expands to a full cycle basis using an ear construction which makes sure the final basis is still fundamental (and partially TR).

The experiments in this paper confirm that GBP can yield very accurate approximations when the loop regions of a Loop-SRG form a fundamental basis and that these approximations can be further improved by choosing a fundamental basis that is at least partially tree robust. The criteria proposed in this paper also lead to approximations that are comparable to IJGP run with a much higher $iBound$ (and therefore much higher space and computational complexity).

These findings open the door for much future work. Rather than simply finding a TR subgraph, as the method in Section 7 does, it would be preferable to have an algorithm that searches for TR bases in a graph or at the very least identifies when a graph does not admit a TR basis. The recommendations in this paper are also purely structural in nature. A natural extension would be to incorporate interaction strengths in the search for suitable loop regions.

The current paper considered pairwise interactions only. A natural extension is to consider factor graphs or more generally region graphs. We argue that we should be looking for a maximal collection of subsets $\{C_i\}$ of outer regions (or factors) that have the property that there exists an ordering $\pi$ for which $C_{\pi(i)} \setminus \{C_{\pi(1)} \cup \cdots \cup C_{\pi(i-1)}\} \neq \emptyset$. In other words, the subsets can be ordered so that subset $C_{\pi(i)}$ has some element that does not appear in any subset preceding it in the ordering. This definition is identical to the one for a fundamental cycle basis (see definition 6) and will guarantee through the reduction rules of [31] that the region graph (factor graph) can be decomposed into independent variable nodes, establishing non-singularity. A next step would then be to define tree-robustness as a collection of region-subsets that can still be decomposed along some ordering if we do not allow certain elements that correspond to the regions of any embedded junction tree to become unique. Again this is very similar to definitions 7 and 8 for TR cycle bases. Whether these generalizations can be captured with mathematical structures as elegant as the theory of cycle spaces (or more generally matroids) remains to be seen and will be left for future research.

### Acknowledgements

This material is based upon work supported by the National Science Foundation under Grant No. 0914783, 0928427, 1018433.

## References

[1] C. Berrou and A. Glavieux. Near optimum error correcting coding and decoding: turbo codes. *IEEE Transactions on Communications*, 44:1261–1271, 1996.

[2] A. Choi and A. Darwiche. An edge deletion semantics for belief propagation and its practical impact on approximation quality. In *Proceedings of the 21st National Conference on Artificial Intelligence (AAAI)*, pages 1107–1114, 2006.

[3] R. Dechter, K. Kask, and R. Mateescu. Iterative join-graph propagation. In *Proceedings of the Eighteenth Conference on Uncertainty in Artificial Intelligence (UAI)*, 2002.

[4] G. Elidan, I. McGraw, and D. Koller. Residual belief propagation: Informed scheduling for asynchronous message passing. In *Proceedings, UAI*, July 2006.

[5] B.J. Frey. *Graphical models for machine learning and digital communication*. MIT Press, 1998.

[6] Y. Gao. The degree distribution of random k-trees. *Theor. Comput. Sci.*, 410:688–695, 2009.

[7] A. Globerson and T. Jaakkola. Convergent propagation algorithms via oriented trees. In *Uncertainty in Artificial Intelligence (UAI)*, 2007.

[8] T. Hazan and A. Shashua. Convergent message-passing algorithms for inference over general graphs with convex free energy. In *The 24th Conference on Uncertainty in Artificial Intelligence (UAI)*, 2008.

[9] T. Heskes. Stable fixed points of loopy belief propagation are minima of the Bethe free energy. In *Neural Information Processing Systems*, volume 15, Vancouver, CA, 2003.

[10] T. Heskes. Convexity arguments for efficient minimization of the Bethe and Kikuchi free energies. *Journal of Machine Learning Research*, 26(153-190), 2006.

[11] Alexander T. Ihler, John W. Fisher, and Alan S. Willsky. Message errors in belief propagation. In Lawrence K. Saul, Yair Weiss, and Léon Bottou, editors, *Neural Information Processing Systems 17*, pages 609–616. MIT Press, Cambridge, MA, 2005.

[12] Mooij J, M and H.J. Kappen. Sufficient conditions for convergence of loopy belief propagation. In *Conference on Uncertainty in Artificial Intelligence (UAI)*, 2005.

[13] T. Kavitha, C. Liebchen, K. Mehlhorn, D. Michail, R. Rizzi, T. Ueckerdt, and K.A. Zweig. Cycle bases in graphs characterization, algorithms, complexity, and applications. *Computer Science Review*, 3(4):199–243, 2009.

[14] R. Kikuchi. A theory of cooperative phenomena. *Physical Review*, 81(6):988–1003, 1951.

[15] F.R. Kschischang, B. Frey, and H.A. Loeliger. Factor graphs and the sum-product algorithm. *IEEE Transactions on Information Theory*, 47(2):498–519, 2001.

[16] D.J.C. MacKay and R.M. Neal. Near shannon limit performance of low density parity check codes. *Electronics Letters*, 32:1645–1646, 1996.

[17] R. McEliece, D. MacKay, and J. Cheng. Turbo decoding as an instance of Pearl's belief propagation algorithm. *IEEE J. Selected Areas in Communication, 1997*, 16:140–152, 1998.

[18] R. McEliece and M. Yildirim. Belief propagation on partially ordered sets. In eds. D. Gilliam and J. Rosenthal, editors, *Mathematical Systems Theory in Biology, Communications, Computation, and Finance*, 1998.

[19] T. Meltzer, A. Globerson, and Y. Weiss. Convergent message passing algorithms - a unifying view. In *Conference Annual Conference on Uncertainty in Artificial Intelligence (UAI)*, pages 393–401, 2009.

[20] T. Morita. Consistent relations in the method of reducibility in the cluster variation method. *Journal Statistical Physics*, 34:319–328, 1984.

[21] T. Morita. Formal structure of the cluster variation method. *Progress of Theoretical Physics supplement*, 115:27–39, 1994.

[22] K. Murphy, Y. Weiss, and M. Jordan. Loopy belief propagation for approximate inference: An empirical study. In *Proc. of the Conf. on Uncertainty in Artificial Intelligence*, 1999.

[23] Payam Pakzad and Venkat Anantharam. Estimation and marginalization using kikuchi approximation methods. *Neural Computation*, 17:1836–1873, 2005.

[24] Judea Pearl. *Probabilistic Reasoning in Intelligent Systems: Networks of Plausible Inference*. Morgan Kaufmann Publishers, San Mateo, California, 1988.

[25] D. Sontag and T. Jaakkola. New outer bounds on the marginal polytope. In *Neural Information Processing Systems*, 2007.

[26] M.J. Wainwright, T. Jaakkola, and A.S. Willsky. Tree-based reparameterization for approximate estimation on loopy graphs. In *Advances Neural Information Processing Systems*, volume 14, vancouver, Canada, 2001.

[27] M.J. Wainwright, T. Jaakkola, and A.S. Willsky. A new class of upper bounds on the log partition function. In *Proc. of the Conf. on Uncertainty in Artificial Intelligence*, Edmonton, CA, 2002.

[28] M.J. Wainwright and M.I. Jordan. Semidefinite relaxations for approximate inference on graphs with cycles. In *Neural Information Processing Systems*, 2004. Rep. No. UCB/CSD-3-1226.

[29] Y. Weiss. Interpreting images by propagation Bayesian beliefs. In *Neural Information Processing Systems*, pages 908–914, 1996.

[30] M. Welling. On the choice of regions for generalized belief propagation. In *Proc. of the Conf. on Uncertainty in Artificial Intelligence*, pages 585–592, 2004.

[31] M. Welling, Tom Minka, and Yee Whye Teh. Structured region graphs: Morphing EP into GBP. In *Proc. of the Conf. on Uncertainty in Artificial Intelligence*, pages 607–614, 2005.

[32] M. Welling and Y.W. Teh. Belief optimization for binary networks: a stable alternative to loopy belief propagation. In *Proc. of the Conf. on Uncertainty in Artificial Intelligence*, pages 554–561, 2001.

[33] J.S. Yedidia, W. Freeman, and Y. Weiss. Bethe free energy, kikuchi approximations, and belief propagation algorithms. Technical report, MERL, 2001. Technical Report TR-2001-16.

[34] J.S. Yedidia, W. Freeman, and Y. Weiss. Constructing free energy approximations and generalized belief propagation algorithms. Technical report, MERL, 2002. Technical Report TR-2002-35.

[35] A.L. Yuille. CCCP algorithms to minimize the Bethe and Kikuchi free energies: Convergent alternatives to belief propagation. *Neural Computation*, 14(7):1691–1722, 2002.